\documentclass[preprint,11pt,numbers]{elsarticle}

\usepackage{amssymb}
\usepackage{amsmath}
\usepackage{comment}
\usepackage{float}
\usepackage[margin=1in]{geometry}
\usepackage{subcaption}
\usepackage{tikz} 
\usepackage{graphicx}
\usepackage{pgffor} 
\usepackage{tabularx}
\usepackage{booktabs}
\usepackage{bm}
\usepackage[ruled,vlined]{algorithm2e}
\journal{Computers, Environment and Urban Systems}
\begin{document}
\begin{frontmatter}

\title{Evaluating Few-Shot Temporal Reasoning of LLMs for Human Activity Prediction in Smart Environments
}

\author[1]{Maral Doctorarastoo}
\ead{mdoctora@andrew.cmu.edu}
\author[1]{Katherine A. Flanigan\corref{cor1}}\ead{kflaniga@andrew.cmu.edu}
\author[1]{Mario Berg\'es\corref{cor2}}
\ead{mberges@andrew.cmu.edu}
\author[2]{Christopher McComb}
\ead{ccm@andrew.cmu.edu}
\cortext[cor1]{Corresponding author at: 5000 Forbes Ave, Pittsburgh, PA, 15213, USA. 
E-mail address: \texttt{kflaniga@andrew.cmu.edu} (K.A. Flanigan)}
\cortext[cor2]{Mario Berg\'es holds concurrent appointments at Carnegie Mellon University (CMU) and as an Amazon Scholar. This manuscript describes work at CMU and is not associated with Amazon.}

\affiliation[1]{organization={Department of Civil \& Environmental Engineering, Carnegie Mellon University},
            addressline={5000 Forbes Ave}, 
            city={Pittsburgh},
            state={PA},
            postcode={15213}, 
            country={USA}
            }

\affiliation[2]{organization={Department of Mechanical Engineering, Carnegie Mellon University},
            addressline={5000 Forbes Ave}, 
            city={Pittsburgh},
            state={PA},
            postcode={15213}, 
            country={USA}}

\begin{abstract}

Anticipating human activities and their durations is essential in applications such as smart-home automation, simulation-based architectural and urban design, activity-based transportation system simulation, and human-robot collaboration, where adaptive systems must respond to human activities. Existing data-driven agent-based models—from rule-based to deep learning—struggle in low-data environments, limiting their practicality. This paper investigates whether large language models, pre-trained on broad human knowledge, can fill this gap by reasoning about everyday activities from compact contextual cues. We adopt a retrieval-augmented prompting strategy that integrates four sources of context—\textit{temporal}, \textit{spatial}, \textit{behavioral history}, and \textit{persona}—%\textit{Temporal} context captures information such as time-of-day and weekday patterns; \textit{spatial} context summarizes the built environment; \textit{behavioral history} encodes recent activities and durations; and \textit{persona} represents individual characteristics and routine tendencies derived from prior observations. 
and evaluate it on the CASAS Aruba smart-home dataset. The evaluation spans two complementary tasks: next-activity prediction with duration estimation, and multi-step daily sequence generation, each tested with various numbers of few-shot examples provided in the prompt. Analyzing few-shot effects reveals how much contextual supervision is sufficient to balance data efficiency and predictive accuracy, particularly in low-data environments. Results show that large language models exhibit strong inherent temporal understanding of human behavior: even in zero-shot settings, they produce coherent daily activity predictions, while adding one or two demonstrations further refines duration calibration and categorical accuracy. Beyond a few examples, performance saturates, indicating diminishing returns. Sequence-level evaluation confirms consistent temporal alignment across few-shot conditions. These findings suggest that pre-trained language models can serve as promising temporal reasoners, capturing both recurring routines and context-dependent behavioral variations, thereby strengthening the behavioral modules of agent-based models. %The broader implications extend to smart home automation, personalized health monitoring, human-robot collaboration, and simulation-based urban planning, where modeling human behavior and user-centered design are essential.
\end{abstract}

%% Keywords

\end{frontmatter}



\section{Introduction}
Predicting human activity is a foundational capability for a wide range of intelligent systems. At short time-scales, near real-time activity prediction enables applications such as autonomous driving, where anticipating pedestrian or cyclist actions is critical for safety \cite{chaabane2020looking, eilbrecht2017model, herman2021pedestrian}, and smart building control, where anticipating occupant needs can optimize comfort and energy use \cite{klein2012coordinating, yang2013development}. At longer time-scales, accurate activity prediction supports simulation-based planning for urban design \cite{yan2024opencity, rezvani2024urban}, evacuation strategies \cite{chang2024simulation, islam2022simulation}, infrastructure resilience \cite{fathianpour2023resilient}, and other strategic decision-making processes such as activity-based travel demand generation \cite{nguyen2025large}. In both operational and design contexts, understanding what people will do next—and for how long—forms the basis for adaptive, human-aware systems \cite{gorur2023fabric, heard2022predicting}.

A broad spectrum of computational approaches has been proposed for activity prediction, each suited to different temporal resolutions and complexity. Rule-based or physics-based models are lightweight and interpretable but rely on hand-crafted assumptions that fail in complex or novel contexts \cite{wu2311smart, camara2020pedestrian}. Classical machine learning (ML) models, such as classifiers and sequence models, offer flexibility but require extensive labeled data to generalize \cite{bouchabou2021survey, presotto2023combining}. More recently, deep learning and imitation learning (IL) have been employed for strategic activity planning, showing improved accuracy in rich, multi-modal environments \cite{zeng2018inverse, kravaris2025transferable, zhang2023pedestrian}. However, these models remain highly data-intensive: their performance scales with the availability of large, domain-specific datasets that are often expensive or infeasible to collect, particularly in emerging or rare scenarios \cite{zheng2025spatio, chen2025spatialllm}. As a result, a persistent trade-off exists between predictive accuracy and data availability, limiting the deployment of data-driven models across diverse time-scales and application domains \cite{zhang2023pedestrian, huang2024gpt}.

\begin{figure}[t!]
\centering
  \includegraphics[width=0.65\linewidth]{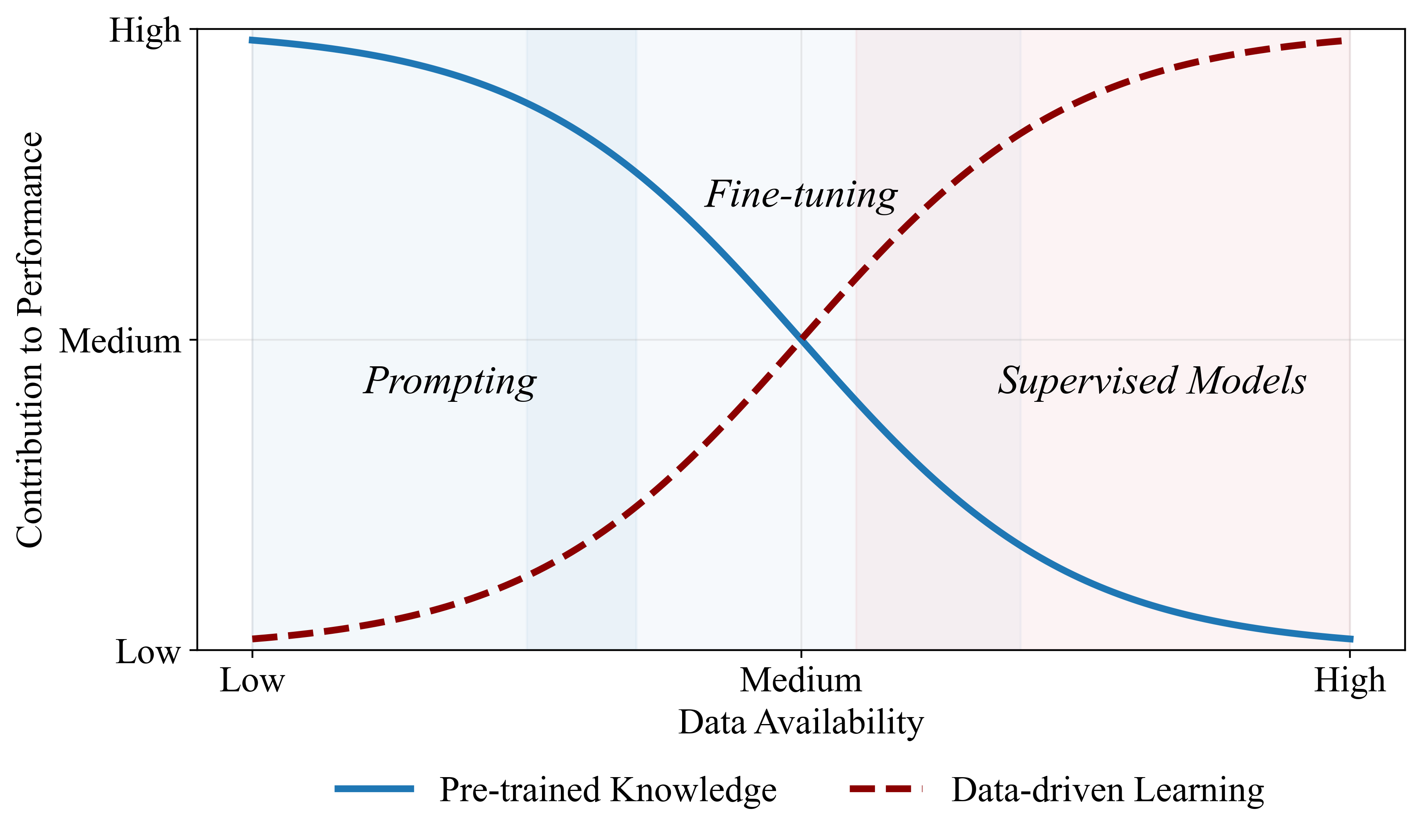}
  \vspace{-0.3cm} \caption{Conceptual illustration of PT and D contribution to model performance under different data regimes.} \vspace{-0.5cm}
  \label{fig:tradeoff}
\end{figure}

Large language models (LLMs) have recently emerged as a promising alternative for human activity prediction. Pre-trained on vast and diverse corpora, LLMs encode rich contextual and commonsense knowledge about human behavior, social norms, and spatial reasoning \cite{martorell2025text, takeyama2024tr}. This pre-training offers a complementary path to accuracy when task-specific data are scarce. Conceptually, the contribution of \textit{pre-trained knowledge (PT)} is highest in low-data regimes but diminishes as more in-domain data become available, while \textit{data-driven learning (D)} contributes little when data are scarce but dominates with large, labeled datasets. Fig.~\ref{fig:tradeoff} schematically illustrates this trade-off suggesting that LLM-based prompting may excel when data are limited, while conventional ML or IL becomes preferable as datasets grow \cite{zheng2025spatio, chen2025spatialllm}.

Recent advances in \textit{LLM-based agent-based modeling (ABM)} have demonstrated LLMs' capacity to predict human plans, reason about physical environments, and model user behavior with minimal supervision \cite{takeyama2024tr, zhao2025navigating}. This emerging paradigm promises to reshape behavioral simulation by enabling agents that reason over natural language, adapt to diverse contexts, and operate without the extensive training data typically required by traditional ABM pipelines. If realized fully, the fusion of LLMs and ABMs could transform applications ranging from smart environments to mobility systems by providing simulators with more flexible, and human-aligned behavioral models. Within this context, activity prediction can be viewed through a situation awareness lens, where human activity sequences act as key observables for assessing and projecting evolving situations over time. Prior work has shown that spatio-temporal activity patterns enable recognition of real-world situations, including group-level caregiving and collaborative operations \cite{sato2016spatio}. Such approaches rely on explicit clustering, probabilistic state models, or rule-based abstractions over activity traces, emphasizing multi-user activity recognition and temporal context as core to situation-aware systems \cite{gravina2022keynote}. In contrast, this work frames next-activity and duration prediction as \emph{situation projection}, evaluating whether LLMs can generalize from compact contextual representations and few in-context examples to predict future human states.

However, translating this conceptual alignment between LLM-based agents and situation projection into reliable system behavior  depends on a series of methodological decisions that influence system performance. These include but are not limited to how contextual information is represented, what forms of supervision are provided, how temporal and physical dynamics are encoded, how agents integrate environmental cues, and how uncertainty in human behavior is handled. Each of these choices can meaningfully shift an LLM agent’s ability to generalize, maintain coherence over time, and align with real-world behavioral patterns. Most existing works focus on community- or population-level behavior rather than individual-level temporal prediction \cite{bougie2025citysim}. Moreover, the performance of LLM-based agents in simulated environments is often assessed through qualitative plausibility or expert judgment, rather than quantitative validation against real-world activity data \cite{hradec2023fables}. Among these design parameters, the number of few-shot demonstrations provided to the model is a central yet understudied factor. Few-shot prompting determines how much behavioral structure the LLM infers from examples versus prior knowledge, shaping its temporal reasoning, duration estimates, and transitions between activities. 

Motivated by this gap, we conduct a data-grounded assessment to examine how LLMs generalize temporal and behavioral patterns from limited contextual evidence and sensor-derived data. Specifically, we focus on \textit{predicting the next human activity and its duration from compact contextual information}, with an emphasis on how few-shot demonstrations effect model performance. The goal is to examine how LLM agents can infer patterns of daily activities from a small number of illustrative examples, providing a lightweight alternative to traditional data-driven ABM methods that require extensive labeled data. %Unlike broader decision-making frameworks that integrate low-level action planning, we isolate and analyze high-level activity planning. 
Our central research question is:
\begin{quote}
\textit{How does the number of few-shot demonstrations included in the prompt affect the accuracy of next-activity as well as multi-step activity prediction?}
\end{quote}

%To investigate this question, we compare prompts that differ only in the number of in-context examples, while keeping all other contextual components and decoding parameters constant. 
The investigation includes two related tasks: \textit{Next–Activity Prediction:} forecasting the immediate next activity and its expected duration; and \textit{Multi-Step Activity Rollout:} predicting an ordered sequence of future activities, each with an estimated duration. These two tasks address complementary needs. Next-activity prediction enables real-time responsiveness in applications such as adaptive smart home control, assistive robotics, and context-aware health monitoring, where predicting the next user action supports timely intervention. In contrast, multi-step activity prediction facilitates analysis and planning in human-environment systems, including behavioral simulation, occupancy forecasting, and the design of human-centered infrastructure systems. To answer these, we compare prompts that differ only in the number of few-shot demonstrations while keeping all other elements and decoding settings fixed. Isolating the marginal contribution of additional demonstrations to performance provides practical guidance for designing prompt-based activity prediction systems, showing when adding more examples meaningfully improves accuracy and when the returns begin to diminish.

To ground this analysis, we conduct an experimental evaluation on the CASAS Aruba smart home dataset \cite{CASAS}, which records daily living activities of a single resident in a sensor-equipped environment. The experiments are designed to provide a data-driven evaluation of few-shot prompting for activity forecasting, offering practical insights into how LLMs can be applied in real-world, data-limited smart environments.

\section{Methodology}
\label{sec:method}

%The study examines how the number of few-shot demonstrations influences both next-activity and multi-step activity prediction. Each prompt integrates temporal, spatial, behavioral history, and persona context while varying only the number of in-context examples. Performance is evaluated using categorical and temporal metrics, including accuracy, F1-score (\(F_{1}\)), mean absolute error (MAE), root mean squared error (RMSE), and symmetric mean absolute percentage error (sMAPE), as well as sequence-level measures such as dynamic time warping (DTW) to capture temporal consistency across predicted activity sequences. Together, these experiments provide an experimental and data-driven evaluation of few-shot prompting for activity forecasting, offering practical insight into how LLMs can be applied in real-world, data-limited smart environments.

We assess how the number of few-shot demonstrations influences LLM performance across two tasks: \textit{next–activity prediction} with duration, and \textit{multi-step sequence rollout}. A \textit{demonstration} is an in-context exemplar consisting of a retrieved ground-truth instance formatted as an \textbf{(context $\rightarrow$ output)} pair, where the context mirrors the information provided to the model at inference time and the output specifies the next performed activity and its duration. %We evaluate performance along two complementary dimensions: \textit{activity prediction} and \textit{duration prediction}. 
In both tasks, predictions are queried at the boundaries between completed activities. Given the \textit{temporal}, \textit{spatial}, \textit{behavioral history}, and \textit{persona} context at each step, the LLM outputs the label of the expected next activity and its estimated duration. The number of few-shot demonstrations included in the prompt is treated as the experimental variable, while all other configuration choices such as model, retrieval method, decoding settings, and thresholds are held constant. Performance is evaluated using categorical and temporal metrics, including accuracy, F1-score (\(F_{1}\)), precision, recall, mean absolute error (MAE), and root mean squared error (RMSE), as well as sequence-level measures such as dynamic time warping (DTW) to capture temporal consistency across predicted activity sequences.

%We examine how the number of few-shot demonstrations influences both next-activity and multi-step activity prediction. 

\subsection{Prompting Framework}

\begin{figure}[t]
  \centering
  \includegraphics[width=0.82\columnwidth]{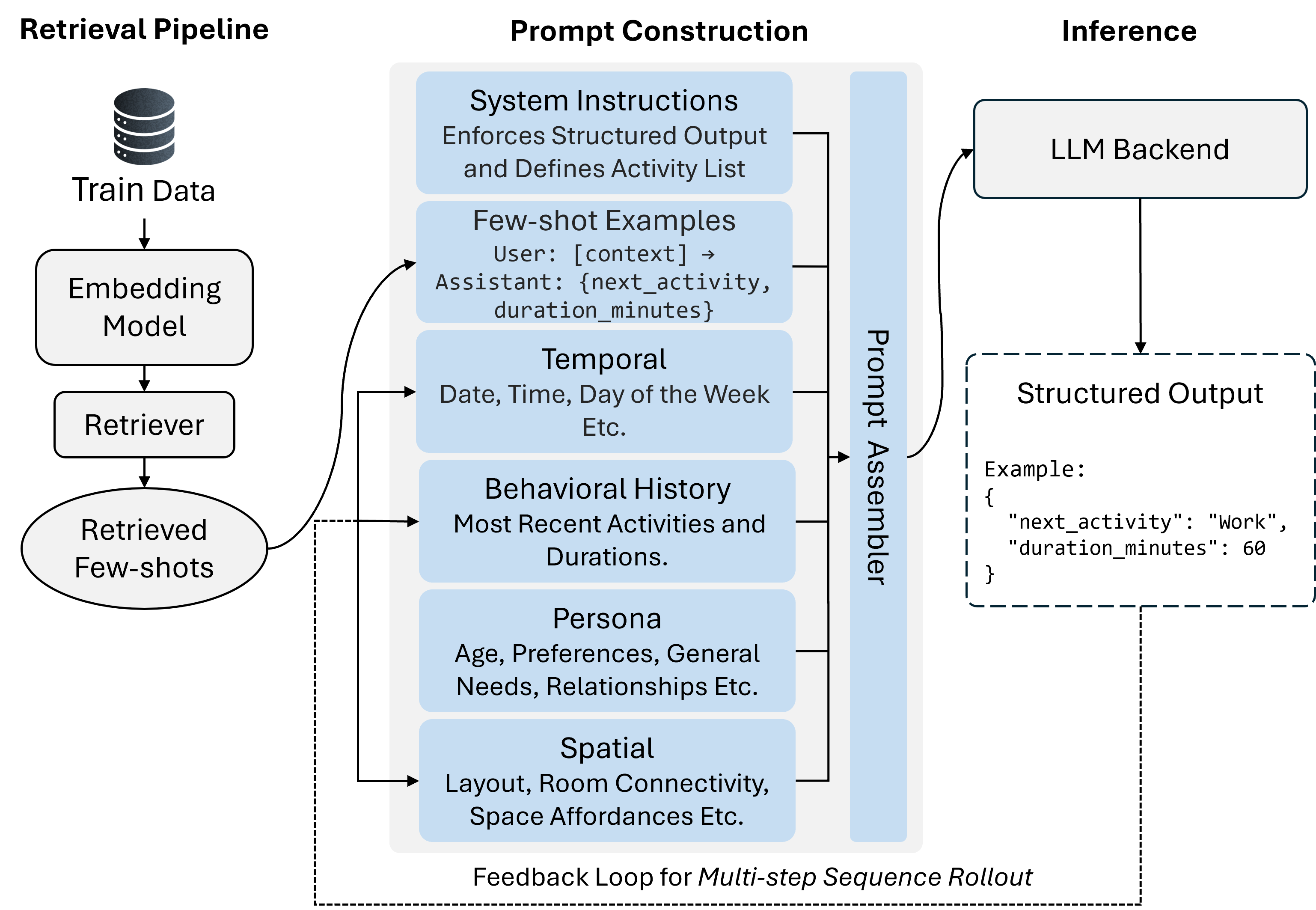}
  \caption{Overview of the prompting framework, comprising three interconnected stages: \textit{retrieval pipeline}, \textit{prompt construction}, and \textit{LLM inference}. Arrows indicate data flow and feedback connections between modules.} \vspace{-0.5cm}
  \label{fig:prompt_framework}
\end{figure}

Fig.~\ref{fig:prompt_framework} illustrates the prompting framework, comprising three interconnected stages: \textit{retrieval pipeline}, \textit{prompt construction}, and \textit{inference}. In the data and retrieval, or \textit{retrieval pipeline} stage, training examples are encoded into a dense vector space using an embedding model capturing temporal and behavioral similarity. Each instance is represented by its day of week, local time, and a compact summary of the most recent activity-duration pairs. A retriever module then selects few-shot demonstrations using maximal marginal relevance (MMR), ensuring that retrieved examples are both highly relevant to the query and mutually diverse. Therefore, instead of accessing the entire training set, the model receives only a few relevant examples resembling past experiences. It then uses these examples to reason about the current situation, much like a person recalling similar past events to guide a decision.

The output of this process is a compact set of structured prompt–response pairs, each formatted in JSON to explicitly represent an input prompt (recent activities, time, persona, etc.) and its corresponding ground-truth output (next activity and its duration). The number of few-shot demonstrations ($N$), is determined by the experimental condition and implicitly conveys the task ontology, enabling the model to infer likely temporal transitions and activity patterns from a small number of prior observations. The \textit{prompt construction} stage integrates the retrieved few-shot demonstrations with multiple layers of contextual information to form a unified natural-language prompt. First a system instruction and an explicit list of allowed activities (“INDEX : NAME”) define the task schema and enforce a deterministic JSON output format with the fields \texttt{next\_activity} and \texttt{duration\_minutes}. Each demonstration, consisting of a prior prompt–response exchange with its contextual input and structured JSON output, is included in the prompt to provide in-context examples of the desired reasoning and response format. This constraint is enforced through structured decoding, which restricts the output to valid JSON format and rejects unstructured text generations.

Four categories of contextual information guide the model’s reasoning: \textit{temporal}, \textit{behavioral history}, \textit{persona}, and \textit{spatial}. The \textit{temporal context} encodes information such as the date, day of week, and time of day, supporting the reasoning about periodic and daily patterns. The \textit{behavioral history} enumerates the most recent activities and their durations in a concise, ordered format, allowing the model to maintain short-term behavioral memory. In the \textit{multi-step activity rollout} setting, this list dynamically updates after each predicted activity to serve as input for the next step. A \textit{persona statement} summarizes individual characteristics and routine tendencies derived from prior observations, expressed in natural language as soft hints. A \textit{spatial description} outlines key aspects of the physical environment, e.g., room connectivity and main functional areas, promoting spatial coherence without enforcing rigid constraints. These elements are combined in the prompt assembler, producing the final structured input for inference.

In the final stage, the assembled prompt is processed by the LLM backend, which leverages its pre-trained knowledge to infer the most likely next activity and its expected duration. The model outputs a structured JSON object, serving either as the final prediction or as input for iterative forecasting. During \textit{multi-step activity rollouts}, each generated output is fed back into the system to update the \textit{behavioral history} and \textit{temporal} context, enabling sequential prediction as the simulated daily activity plan unfolds. This structured yet compact prompt formulation enables the model to reason about human activity sequences from minimal supervision, leveraging temporal regularities, behavioral continuity, and a small number of representative demonstrations.

\subsection{Evaluation Metrics}
\label{sec:metrics}

We evaluate two targets per instance $i$: the next activity label $y_i \in \mathcal{C}$ and its duration $t_i \in \mathbb{R}_{>0}$, with model outputs $\hat{y}_i$ and $\hat{t}_i$. For classification, we report $F_{1}$, precision, and recall scores aggregated as micro-, macro-, and weighted averages to reflect class imbalance among daily activities. Micro-$F_{1}$ is computed using global true positive (TP), false positive (FP), and false negative (FN), and is numerically identical to fraction of samples where the predicted label is correct. In this single-label multi-class setting, micro-precision, micro-recall, and micro-$F_1$ are identical and reduce to overall classification \textit{accuracy}. Macro-averaged precision, recall, and $F_1$ are unweighted across classes, whereas weighted-averaged metrics weight each class by its support.

For duration quality, with error $e_i=\hat{t}_i-t_i$, we report absolute and quadratic losses,
\[
\mathrm{MAE}=\frac{1}{n}\sum_{i=1}^{n}|e_i|, \qquad
\mathrm{RMSE}=\sqrt{\frac{1}{n}\sum_{i=1}^{n}e_i^2}.
\]

To jointly evaluate categorical and temporal performance, we report two composite measures. The \textit{Joint Success@$T$m} metric counts instances with both a correct activity label and a duration prediction within $T$ minutes of the ground truth,
\[
\mathbf{1}\!\left\{\,\hat{y}_i=y_i \ \land\  |\hat{t}_i-t_i|\le 10\,\right\}.
\]
\begin{comment}
    
The \emph{Penalized MSE} integrates duration error and classification penalty into a single loss,
\[
\mathcal{L}_i=(\hat{t}_i-t_i)^2+\lambda\,\mathbf{1}\{\hat{y}_i\neq y_i\},
\]
where $\lambda$ is set to the global mean squared duration from the training set, placing misclassification penalties on the same $\mathrm{min}^2$ scale as duration errors.
\end{comment}

For multi-step daily rollouts, we assess sequence-level temporal fidelity using DTW. Each day is expanded into a one-minute categorical timeline; predicted and ground-truth sequences are aligned with a $0/1$ mismatch cost (0 if labels match, 1 otherwise). We report the raw DTW (in minutes), representing the minimum number of minutes of disagreement after optimal warping, and a normalized version,
\[
\mathrm{DTW}_{\mathrm{norm}}=\frac{\mathrm{DTW}_{\mathrm{raw}}}{T_{\mathrm{day}}},
\]
where $T_{\mathrm{day}}$ denotes the total evaluated duration of the day, allowing comparison across days. All metrics are computed on a held-out test split.

\section{Case Study: CASAS Aruba}
\label{sec:aruba}

\subsection{Dataset Overview}

\begin{figure}[t!]
\centering
\includegraphics[width=0.55\linewidth]{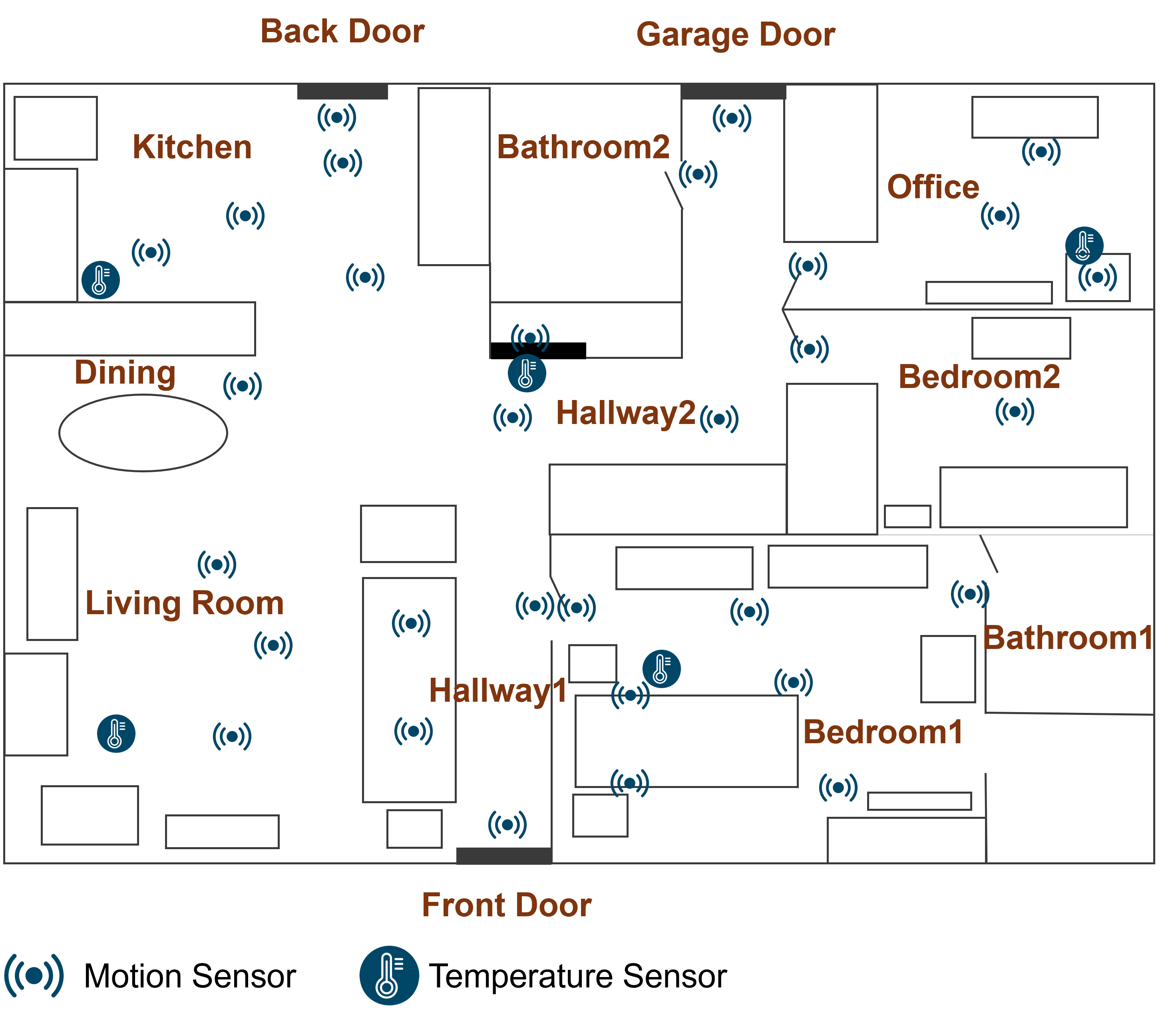}
\vspace{-0.2cm} \caption{CASAS Aruba floor plan and sensor layout.} \vspace{-0.4cm}
\label{fig:floor_plan}
\end{figure}

We conduct a case study on the CASAS Aruba smart home dataset. % to ground the methodology %in a real deployment context.
 The Aruba environment is instrumented with ambient binary motion and door sensors %(with identifiers \texttt{M\#\#\#} and \texttt{D\#\#\#}, respectively)
, while temperature sensors %(\texttt{T\#\#\#}) 
are present but not used for activity recognition because they do not capture movement. Motion and door sensors report ON/OFF or OPEN/CLOSE states, respectively.
%, and sensor numbering is arbitrary with respect to layout. 
The dataset description for Aruba reports a single-resident household, comprising 39 sensors and 11 activity classes over 219 days. Fig.~\ref{fig:floor_plan} illustrates the Aruba floor plan and sensor locations.

%\begin{table}[t]
%\caption{Key Statistics for the CASAS Aruba Dataset}
%\begin{center}
%\begin{tabular}{|c|c|c|c|}
%\hline
%\textbf{Monitoring Days} & \textbf{Sensors} & \textbf{Residents} %& \textbf{Unique Activities} \\
%\hline
%219 & 39 & 1 & 11 \\
%\hline
%\end{tabular}
%\label{tab:data_stats_aruba}
%\end{center}
%\end{table}

\subsection{Experimental Setup on Aruba}

We instantiate the pipeline introduced in Section~\ref{sec:method} on the Aruba home. % without adding dataset-specific heuristics. 
The raw event stream is filtered to retain lines containing \texttt{begin} or \texttt{end} markers corresponding to activity boundaries. Each \emph{begin-end} pair of the same label is matched sequentially in a single pass to form completed intervals, while malformed or unmatched sequences are excluded. In addition, \texttt{being\_outside} intervals are explicitly constructed by pairing \texttt{leave\_home} and \texttt{enter\_home} events, ensuring that out-of-home periods are represented consistently with other activities. This alignment allows the full activity set to remain uniform in structure and semantics. %, supporting coherent transition modeling across all classes.

\begin{figure}[b!]
\centering
 \vspace{-0.0cm}
  \includegraphics[width=0.7\linewidth]{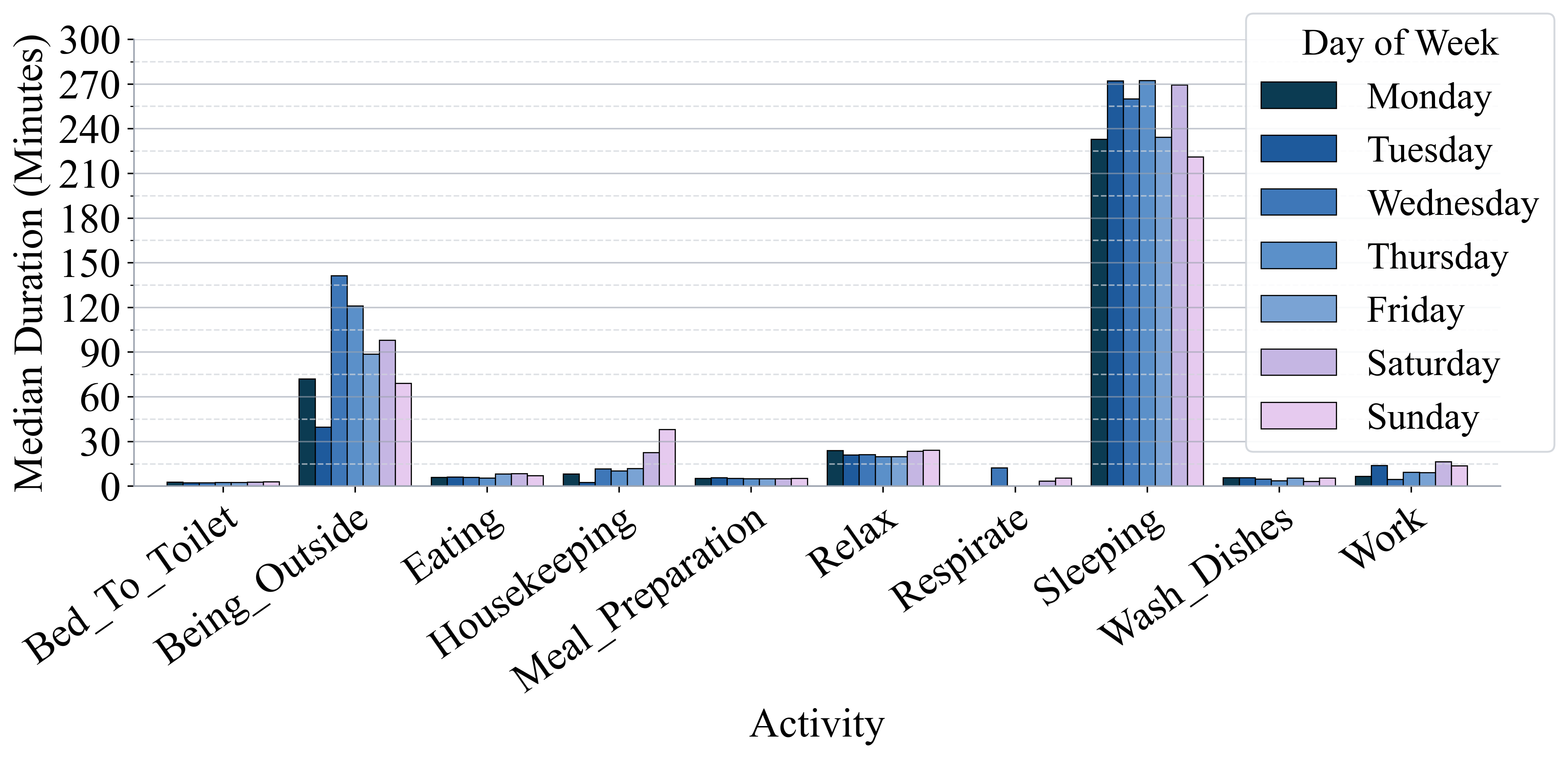}
  \vspace{-0.2cm} \caption{Median duration of activities across days of the week.}
 \vspace{-0.1cm}
  \label{fig:median_duration_bar}
\end{figure}

\begin{figure}[t!]
  \centering
  \vspace{-0.2cm}

  \begin{subfigure}[t]{0.48\linewidth}
    \centering
    \includegraphics[width=\linewidth]{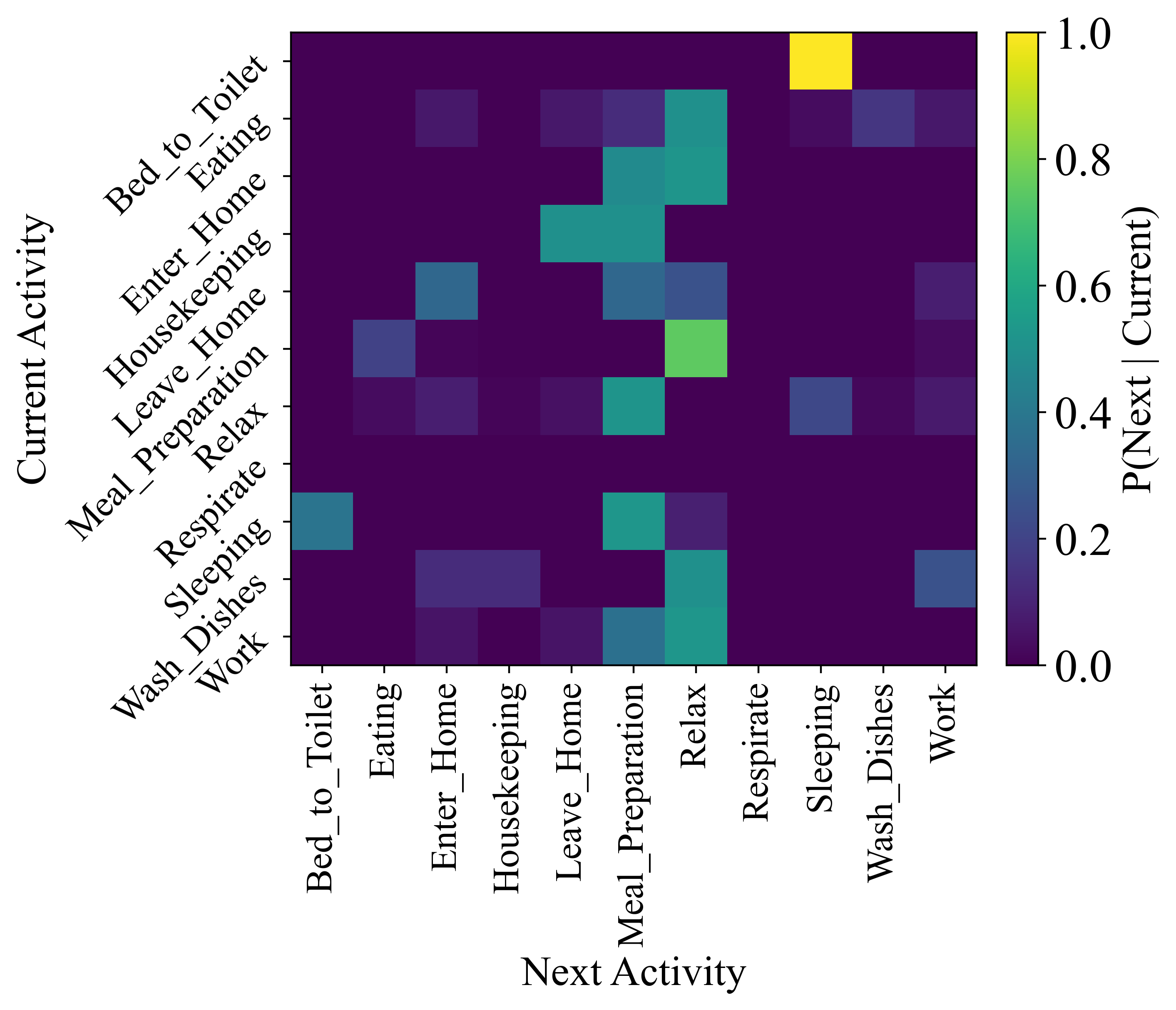}
    \caption{Representative weekday (Thursday).}
    \label{fig:transition_thursday}
  \end{subfigure}
  \hfill
  \begin{subfigure}[t]{0.48\linewidth}
    \centering
    \includegraphics[width=\linewidth]{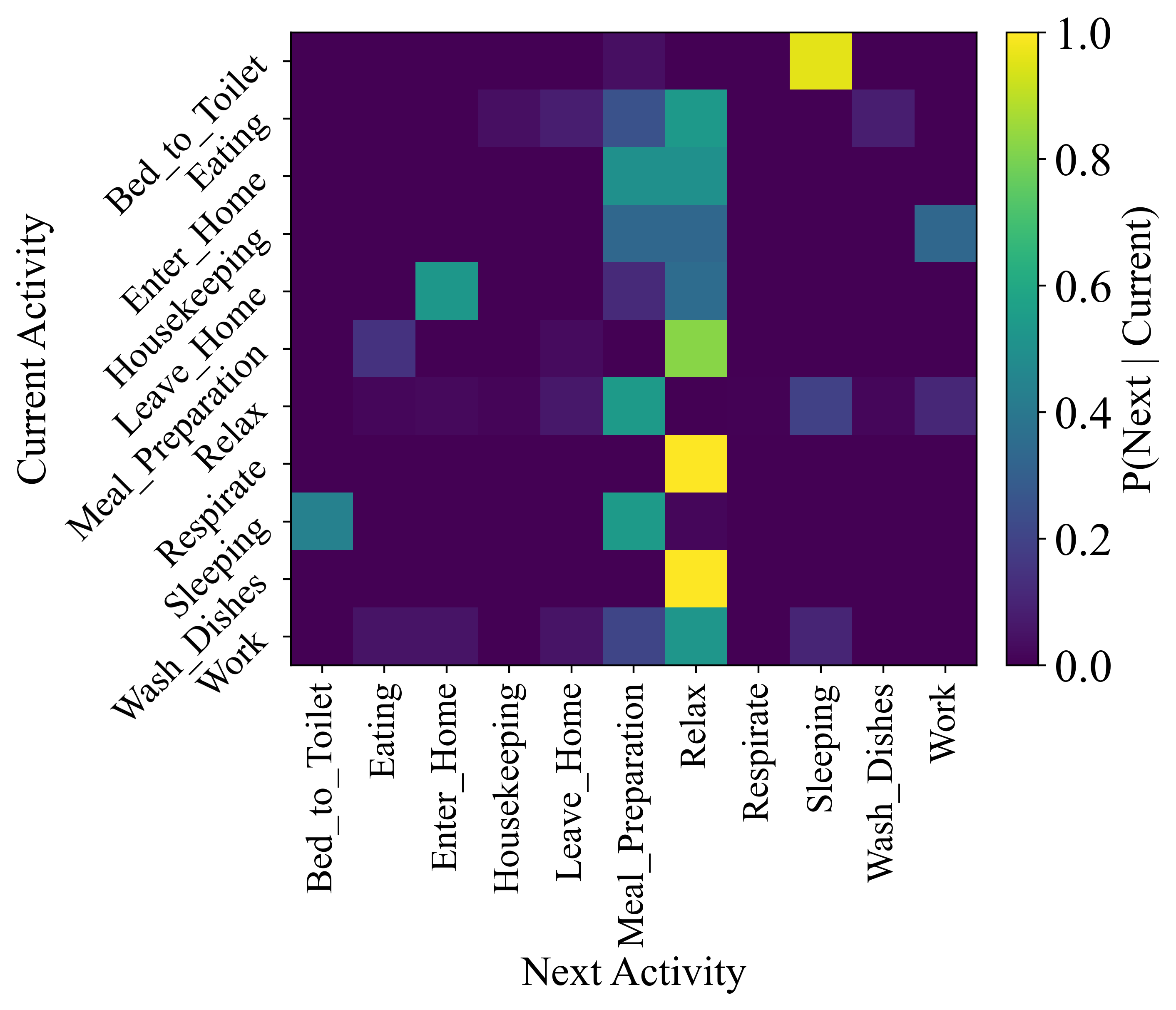}
    \caption{Representative weekend day (Sunday).}
    \label{fig:transition_sunday}
  \end{subfigure}

  \vspace{-0.3cm}
  \caption{Activity transition matrices for weekday and weekend routines. Color intensity indicates the probability of transitioning from the activity on the Y-axis (current activity) to the activity on the X-axis (next activity).}
  \label{fig:activity_transition_comparison}
\end{figure}

Each retained interval includes the activity label, start and end timestamps, duration in minutes, and day of the week. The dataset is split chronologically (80\% training, 20\% evaluation) to preserve temporal order and prevent information leakage. From the training portion, two lightweight priors are estimated for inference: next-activity transition probabilities conditioned on the previous activity and day of week, and median activity durations by activity and day of week (Fig.~\ref{fig:median_duration_bar}). Transition matrices are further stratified into 15-minute time slots to capture fine-grained temporal variation throughout the day (see Fig.~\ref{fig:transition_thursday} for a representative weekday and Fig.~\ref{fig:transition_sunday} for a weekend example). These slot-level transition priors are also used within a time-aware Markov baseline, which provides a non-LLM reference for comparison and contextual evaluation. The baseline selects the next activity by sampling from the learned transition probabilities for the last finished activity, day of week, and 15-minute slot, with fallbacks when a slot-specific matrix is unavailable (day-level, then overall). The predicted duration is assigned using the training median for that activity on the same day of week, falling back to the activity’s global median when needed.

%Each retained interval includes the activity label, start and end timestamps, duration in minutes, and day of the week. The dataset is split chronologically (80\% training, 20\% evaluation) to preserve temporal order and prevent information leakage. From the training portion, two lightweight priors are estimated for inference: next-activity transition probabilities conditioned on the previous activity and day of week, and median activity durations by activity and day of week. Transition matrices are further stratified into 15-minute time slots to capture fine-grained temporal variation throughout the day. These slot-level transition priors are also used within a time-aware Markov baseline, which provides a non-LLM reference for comparison and contextual evaluation. The baseline selects the next activity by sampling from the learned transition probabilities for the last finished activity, day of week, and 15-minute slot, with fallbacks when a slot-specific matrix is unavailable (day-level, then overall). The predicted duration is assigned using the training median for that activity on the same day of week, falling back to the activity’s global median when needed.

Each test instance supplied to the LLM includes the three most recent completed activities with their durations, natural-language time context (day of week and local time), and a concise persona representing a single-resident household. A short prose description of room adjacency is provided solely as context and does not impose hard spatial constraints. %Prompts require a structured JSON reply specifying the next activity and its duration in minutes, enabling deterministic parsing and metric computation.

\section{Model and Implementation Details}
\label{sec:model_details}

%The implementation settings for all experiments are summarized below. 
All experiments employed the \texttt{gpt-4o-mini} model through the OpenAI API in JSON mode without fine-tuning. The prompt history window was fixed at three prior activities. The temperature parameter controlled sampling stochasticity, with \texttt{Temperature=0.0} yielding deterministic outputs. The \texttt{all-MiniLM-L6-v2} embedding model was used to encode temporal and behavioral similarity during retrieval. The same model weights and decoding configuration were used across all runs to ensure consistent comparisons.

\begin{comment}

\end{comment}

\section{Results}
\label{sec:results}
\subsection{Next-Activity Prediction}
\label{sec:results_next}

This section analyzes the \textit{Next-Activity Prediction} task. As shown in Fig.~\ref{fig:fewshot_f1_variants}, \(F_{1}\) scores rise when adding the first one or two demonstrations and plateau thereafter.  
Accuracy increases from \(0.494\) at \(N{=}0\) to a maximum of \(0.528\) at \(N{=}2\), while macro-\(F_{1}\) score reaches its peak at \(N{=}1\). Beyond three demonstrations, accuracy gains are minimal, suggesting that the model rapidly generalizes temporal and contextual patterns from a small number of exemplars.

\subsubsection{Duration Prediction Quality}

The effect of few-shot prompting on duration estimation is more nuanced.  
Absolute errors (MAE and RMSE) improve sharply from the zero-shot baseline at \(N{=}1\), dropping from \(39.6\) to \(34.9\) minutes in MAE and from \(87.9\) to \(79.2\) in RMSE.  
However, the error stabilizes or slightly increases for \(N{>}1\), indicating that a single example is sufficient for the model to calibrate its sense of temporal scale.

%\begin{figure}[b!]
%\centering
%  \includegraphics[width=0.9\linewidth]{figures/fewshot_duration_errors.png}
%  \vspace{-0.3cm} \caption{Effect of few-shot demonstrations on duration estimation.  
%  }\label{fig:fewshot_duration_errors}
%\end{figure}

\subsubsection{Joint Activity–Duration Performance}

\begin{figure}[tb!]
\centering
  \includegraphics[width=0.60\linewidth]{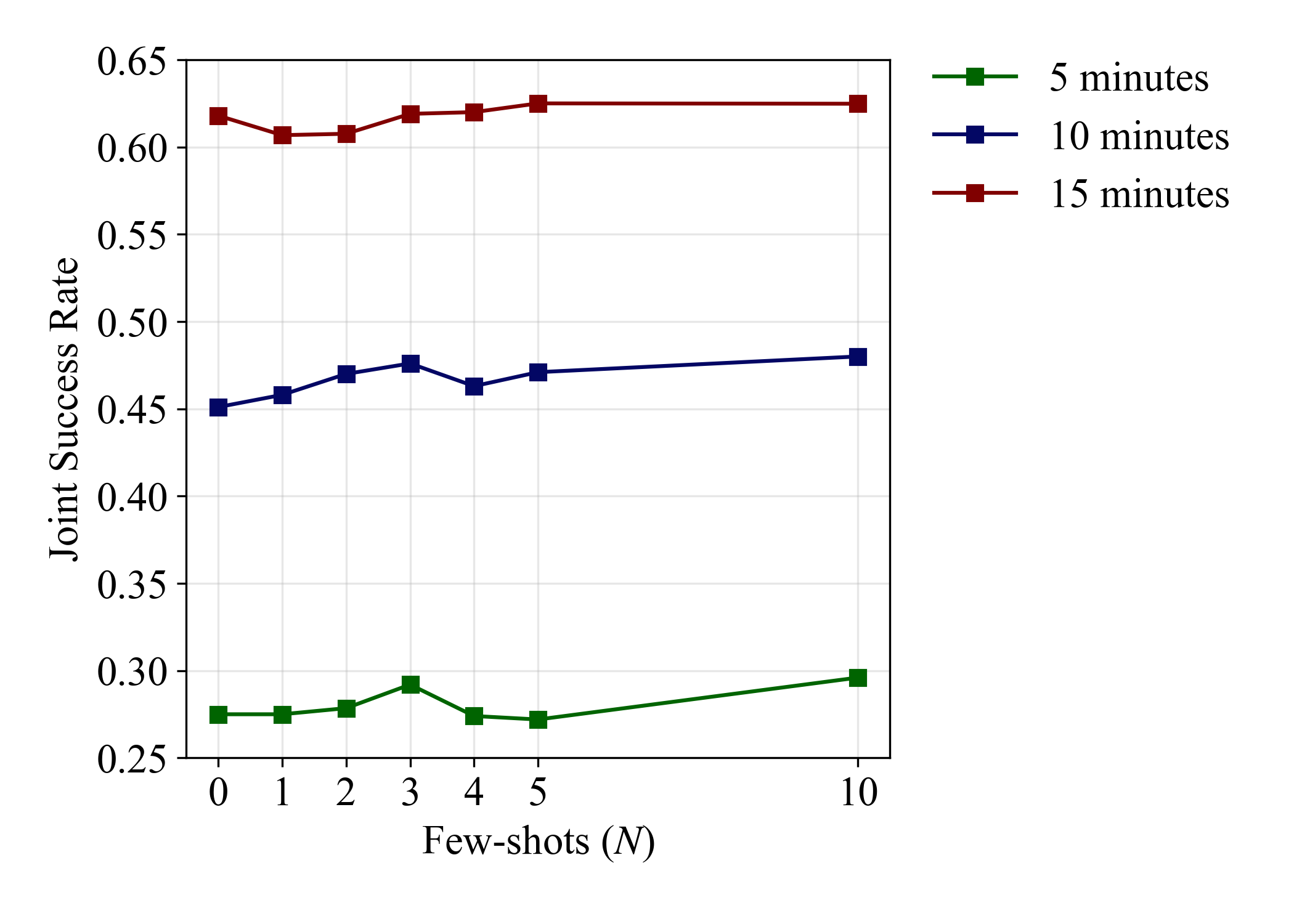} \vspace{-0.5cm}
  \caption{Joint success rates vs.\ few-shot count \(N\).}
  \label{fig:fewshot_joint_metrics}
\end{figure}

Joint metrics, which require both a correct label and an accurate duration, continue improving as \(N\) increases.  
The joint success rate rises from \(0.275\), \(0.451\), and \(0.618\) at \(N{=}0\) to \(0.296\), \(0.480\), and \(0.625\) at \(N{=}10\) for 5-, 10-, and 15-minute tolerances, respectively, as shown in Fig.~\ref{fig:fewshot_joint_metrics}.

%The penalized end-to-end MSE, which integrates both classification and timing errors, decreases by nearly half (from \(\sim3213\) to \(\sim1625\)) over the same range.  
%Fig.~\ref{fig:fewshot_joint_metrics} illustrates this consistent downward trend, suggesting that larger few-shot sets help the model align temporal structure with activity semantics.

\subsubsection{Interpretation}

Table~\ref{tab:fewshot_summary} summarizes the key metrics at representative few-shot settings, and together with the few-shot scaling curve, reveals three distinct performance regimes:

%Table~\ref{tab:fewshot_summary} and few-shot scaling curve reveal three distinct performance regimes:

\begin{itemize}
    \item \textit{Zero to one shot:}  
    The largest reduction in absolute duration error, showing that a single example strongly calibrates temporal reasoning.
    
    \item \textit{One to three shots:}  
    Peak label accuracy and \(F_{1}\); beyond this point, accuracy gains saturate.
    
    \item \textit{Three to ten shots:}  
    Improvements in consistency and joint metrics (lower penalized risk, higher joint success), reflecting more coherent coupling of activity and duration predictions.
\end{itemize}

These regimes suggest practical prompt design strategies: 
few-shot prompting with \(2 \leq N \leq 3\) offers a near-optimal trade-off between accuracy and efficiency for real-time or resource-limited settings, 
whereas \(5 \leq N \leq 10\) is preferable for simulation or planning tasks that require temporal coherence and end-to-end reliability.

\begin{table}[b!]
\centering
\caption{Summary of few-shot next–activity results on Aruba.}
\begin{tabular}{lcccc}
\hline
\textbf{Metric} & \textbf{$N=0$} & \textbf{$N=1$} & \textbf{$N=2$} & \textbf{$N=10$} \\
\hline
Accuracy & 0.494 & 0.519 & \textbf{0.528} & 0.519 \\
Macro-\(F_{1}\) & 0.479 & 0.517 & \textbf{0.526} & 0.517 \\
MAE [min] & 39.6 & \textbf{34.9} & 36.1 & 38.6 \\
RMSE [min] & 87.9 & \textbf{79.2} & 84.1 & 86.2 \\
Joint Success@5  & 0.275 & 0.275 & 0.279 & \textbf{0.296} \\
Joint Success@10 & 0.451 & 0.458 & 0.470 & \textbf{0.480} \\
Joint Success@15 & 0.618 & 0.607 & 0.608 & \textbf{0.625} \\\hline
\end{tabular}
\label{tab:fewshot_summary}
\end{table}

\subsection{Effect on Class Balance and Generalization}

\begin{figure}[h!]
\centering
  \includegraphics[width=0.7\linewidth]{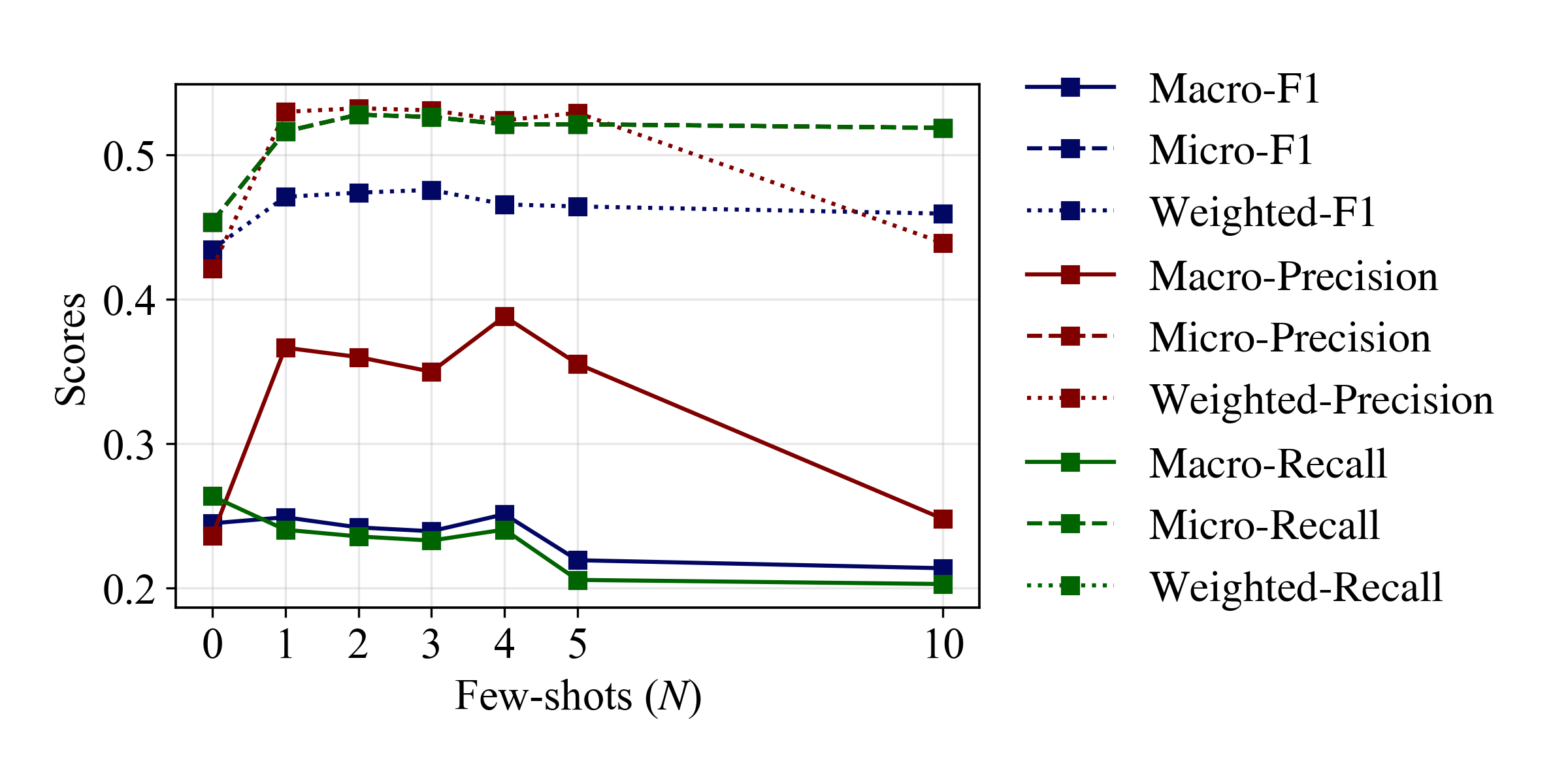} \vspace{-0.5cm}
  \caption{Accuracy, macro- and weighted-\(F_{1}\), precision, and recall vs.\ few-shot count \(N\). } \label{fig:fewshot_f1_variants} \vspace{-0.3cm}
\end{figure}

Fig.~\ref{fig:fewshot_f1_variants} compares accuracy, macro- and weighted-\(F_{1}\), precision, and recall as a function of the number of few-shot demonstrations. The curves reveal several distinct phases of improvement.

\subsubsection{Rapid Early Gain} A sharp performance increase occurs from zero- to two-shot prompting. This regime reflects the model’s transition from relying solely on pre-trained priors to incorporating task-specific behavioral cues. The first few demonstrations dramatically reduce uncertainty in label boundaries and temporal continuity, leading to large jumps in accuracy and weighted-\(F_{1}\). This shows that even a handful of examples is sufficient for the LLM to infer the ontology of daily activities and the general distribution of transitions.

%Performance improves steeply up to two-shot prompting and then stabilizes, showing diminishing returns and improved class balance.

\subsubsection{Stabilization and Saturation} Accuracy increases sharply from \(N{=}0\) to \(N{=}1\) and reaches its peak around \(N{=}2\). Accuracy (and equivalently weighted-recall) increases from \(0.454\) to \(0.516\) at \(N{=}1\) and to \(0.528\) at \(N{=}2\), after which it remains within a narrow band (\(\approx 0.519\text{--}0.526\)). Weighted-\(F_{1}\) shows only marginal improvement through \(N{=}3\) (from \(0.435\) to \(0.476\)) and then gradually declines, indicating that additional demonstrations provide limited benefit and may introduce mild instability.

\subsubsection{Precision--recall Imbalance Across Classes}
Macro-level behavior reveals a different pattern than aggregate metrics. Macro-recall decreases from \(0.264\) at \(N{=}0\) to \(\approx 0.233\) by \(N{=}3\) and continues dropping to \(\approx 0.203\) by \(N{=}10\), indicating that class-balanced coverage worsens as more demonstrations are added. In contrast, macro-precision is comparatively unstable: it rises substantially at \(N{=}1\) (\(0.237 \rightarrow 0.367\)), reaches its highest value at \(N{=}4\) (\(0.388\)), and then drops sharply at \(N{=}10\) (\(0.248\)). Weighted precision follows a similar pattern at the aggregate level, remaining high from \(N{=}1\) through \(N{=}5\) (\(\approx 0.524\text{--}0.532\)) before collapsing at \(N{=}10\) (\(0.439\)). Together, these trends suggest that additional demonstrations primarily increase selectivity (precision) early on, while recall for less frequent classes steadily deteriorates, and high-shot prompting can destabilize precision.

%\subsubsection{Precision-recall trade-off} Precision rises faster than recall as the number of demonstrations increases, showing that the model becomes more confident and selective in its predictions. This asymmetry indicates stronger contextual calibration—fewer false positives—but suggests that additional mechanisms (e.g., retrieval diversity or balanced sampling) might be needed to capture infrequent transitions.

\subsection{Multi-step Sequence Rollout}

To evaluate the temporal alignment and structural fidelity of predicted activity sequences, we compute DTW between predicted and ground-truth daily activity timelines. DTW provides a sequence-level measure of dissimilarity that jointly accounts for both label mismatches and temporal misalignments. Lower DTW values indicate higher similarity between predicted and actual activity patterns. We report results in both raw form (minutes) and normalized form (scaled by the total daily activity duration), and we compare them to a Markov baseline model that uses transition probabilities estimated from the training set.

Normalized DTW values remain consistently low across all few-shot conditions, ranging from $0.12$ to $0.15$, compared to the baseline’s $0.27$. This corresponds to an average daily temporal misalignment of only 10--15\% of total activity time, a more than twofold improvement over Markov baseline predictions. The lowest normalized DTW occurs at $N{=}0$ and $N{=}1$ ($0.12$), suggesting that even without multiple demonstrations the model aligns activity timing through its inherent commonsense priors. Increasing the number of demonstrations to $N{=}5$ or $10$ introduces only marginal fluctuation ($\Delta \mathrm{DTW} < 0.03$), confirming that the model’s inherent commonsense priors capture most temporal regularities without needing extensive few-shot adaptation. Variance slightly increases for higher $N$, implying that additional examples diversify plausible activity sequences rather than improving precision. Raw DTW values show parallel trends, with medians between 142 and 170 minutes across few-shot levels, compared to 309 minutes for the baseline.

\begin{figure*}[b!]
    \centering
    \includegraphics[width=1\textwidth]{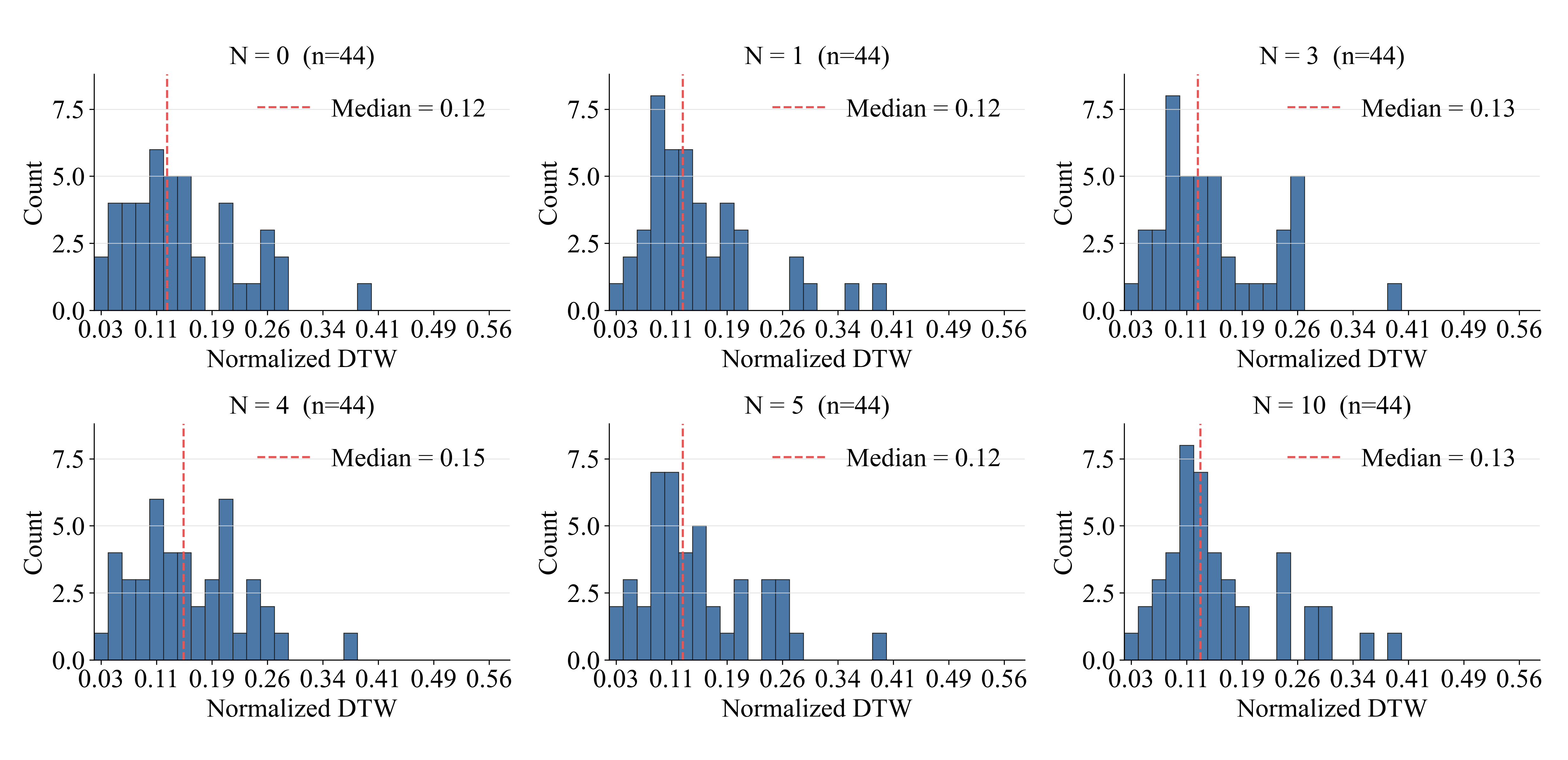}
    \vspace{-0.5cm} \caption{Distribution of normalized DTW scores across few-shot settings ($N{=}0,1,2,3,4,5,10$). Dashed red lines mark medians.} \vspace{-0.4cm}
    \label{fig:dtw_pred_norm_hist}
\end{figure*}

\begin{figure}[t!]
    \centering
    \includegraphics[width=0.55\linewidth]{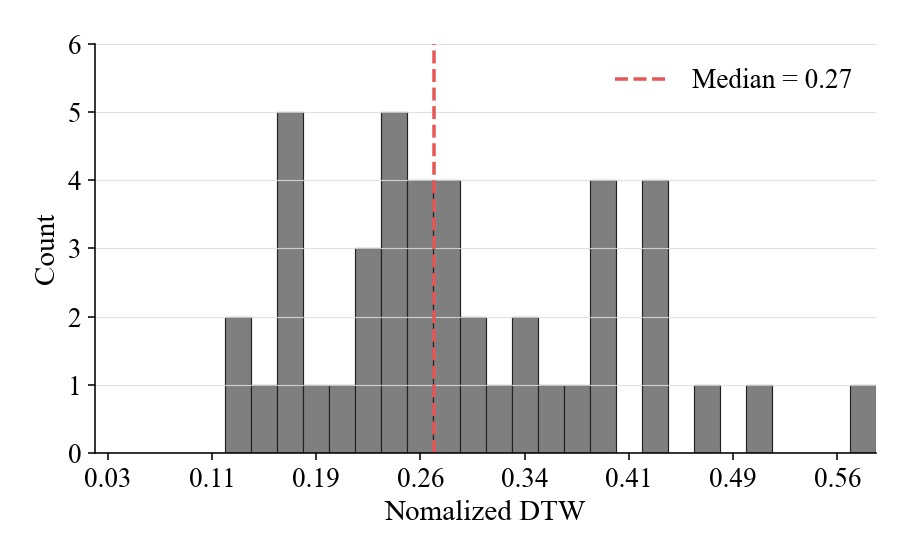} \vspace{-0.4cm}
    \vspace{-0.0cm} \caption{Baseline normalized DTW distribution.}
    \label{fig:dtw_pred_norm_baseline} \vspace{-0.5cm}
\end{figure}

%\begin{enumerate}
%    \item \textbf{LLMs achieve strong temporal coherence.} Both raw and normalized DTW show that even zero-shot predictions align closely with daily behavioral rhythms. The LLM captures routine transitions such as sleep–meal–work cycles without any demonstrations, highlighting robust temporal commonsense reasoning.
%    \item \textbf{Few-shot prompting adds diversity, not accuracy.} The narrow DTW range across $N{=}0$–$10$ confirms that few-shot demonstrations primarily diversify possible trajectories rather than improving alignment quality. Temporal structure is already well-internalized from pre-training.
 %   \item \textbf{Rule-based baselines underperform consistently.} The transition-prior baseline’s DTW medians ($0.27$ normalized, $309$ minutes raw) are roughly double those of all LLM conditions, underscoring the advantage of natural language reasoning over fixed probabilistic rules.
%\end{enumerate}

\subsection{Interpretation and Insights}

\subsubsection{Intrinsic Temporal Reasoning and Stability} Both normalized and raw DTW values demonstrate that the LLM inherently preserves temporal coherence in daily activity sequences. In Fig.~\ref{fig:dtw_pred_norm_hist} we visualize DTW results as histograms to capture the full distribution of sequence-level errors across test days, revealing both central tendency and variability that are critical for assessing the robustness and consistency of multi-step predictions. Across all few-shot settings, normalized DTW values remain consistently low ($0.12$--$0.15$), far below the baseline distribution in Fig.~\ref{fig:dtw_pred_norm_baseline}. Even in the zero-shot condition ($\mathrm{DTW_{norm}}{=}0.12$, $\mathrm{DTW_{raw}}{\approx}145$\,min), predictions align with realistic circadian rhythms—sleeping during night hours, meal preparation around morning or evening, and short relaxation phases in between. The minimal difference between $N{=}0$ and $N{=}10$ (${\Delta}\mathrm{DTW_{norm}}{<}0.03$) indicates that temporal structure is not learned through few-shot demonstrations but retrieved from pre-trained commonsense priors about human daily life.

%\begin{figure}[t!]
%    \centering
%    \includegraphics[width=\linewidth]{figures/dtw_histograms__dtw_pred_norm__median_bars.png}
%    \vspace{-0.5cm} \caption{Median normalized DTW values per few-shot setting. Lower values indicate closer temporal alignment.}
%    \label{fig:dtw_pred_norm_medians}
%\end{figure}

\subsubsection{Few-shot Prompting Enhances Contextual Flexibility rather than precision} Increasing the number of demonstrations from one to ten produces small oscillations in DTW (142–170\, minutes), suggesting that additional few-shots introduce variability in plausible sequences rather than consistent improvement. This pattern implies that the model uses examples to explore alternate but equally valid temporal structures, such as shifting meal or relaxation times, thereby increasing behavioral diversity while maintaining global coherence. In other words, few-shot prompting modulates \textit{when} activities occur, not \textit{whether} they occur.

\subsubsection{Structured Prompting Outperforms Statistical Priors by Over 50\%} The Markov baseline, grounded in empirical transition probabilities, yields median DTW values roughly twice those of all LLM settings (0.27 normalized, 309\,min raw). This contrast underscores the limitation of fixed Markovian assumptions, which fail to capture nonlinear contextual dependencies. In contrast, the LLM integrates temporal, semantic, and contextual cues holistically, yielding smoother daily trajectories and more human-like variability.

\subsubsection{Implications for Generalizable Human Modeling} The results reveal that strong temporal generalization can emerge without task-specific fine-tuning. LLMs encode an implicit model of human chronobiology and behavior scheduling, allowing them to reconstruct realistic daily timelines with minimal contextual supervision. This suggests that prompt-based temporal reasoning could serve as a lightweight alternative to sequence learning in domains where collecting dense, labeled time-series data is infeasible.

%These results demonstrate that large language models inherently reproduce realistic temporal organization of human daily activities. Few-shot prompting provides only marginal refinement, implying that strong temporal generalization is an emergent property of pre-trained linguistic representations rather than a result of in-context adaptation. 
While the findings highlight consistent patterns in how few-shot demonstrations shape LLM-based activity prediction, they are derived from a single household dataset with a specific spatial configuration and behavioral routine. As such, these results should be interpreted as case-specific evidence of underlying mechanisms rather than universal performance claims.

\section{Conclusion}

This work investigated how the number of few-shot demonstrations influence large language models’ ability to predict human activities and their durations. Through systematic experiments on next–activity prediction and sequence–level temporal alignment, we found that LLMs exhibit strong intrinsic reasoning about daily routines even in zero–shot settings. Few–shot prompting provided marginal gains in categorical accuracy and temporal alignment, suggesting that much of the model’s behavioral knowledge stems from its pre-trained commonsense priors rather than from in-context examples. Across both experiments, the model captured the structure of daily activities and their durations with over 50\% improvement in DTW to Markov baselines.

This study focused on one smart–home dataset. Still, broader validation across multi-resident homes, diverse populations, and richer sensor modalities is necessary to confirm whether these patterns generalize to more complex and less structured environments. Future work may extend this approach toward adaptive retrieval, context pruning, or hybrid integration with data-driven reinforcement learning. At a broader level, these findings reinforce the emerging view of pre-trained language models as general cognitive priors for human-centered cyber–physical–social systems. Their ability to infer temporal structure, spatial regularities, and behavioral intent from compact prompts suggests a new paradigm for modeling human dynamics without heavy supervision or extensive labeled data. Such capabilities could inform the design of transparent and adaptive systems for personalized healthcare, energy-efficient buildings, assistive robotics, and simulation-driven urban design.

\section*{Acknowledgment}

This work is supported by the National Science Foundation under Grant \#2425121.
%The preferred spelling of the word ``acknowledgment'' in America is without  an ``e'' after the ``g''. Avoid the stilted expression ``one of us (R. B.  G.) thanks $\ldots$''. Instead, try ``R. B. G. thanks$\ldots$''. Put sponsor acknowledgments in the unnumbered footnote on the first page.

\bibliographystyle{elsarticle-harv}
\bibliography{cas-refs}

\end{document}